\pdfoutput=1

\documentclass[11pt]{article}

\usepackage{EMNLP2023}

\usepackage{times}
\usepackage{latexsym}
\usepackage{algorithm}
\usepackage{algpseudocode}
\usepackage{booktabs}
\usepackage{tabularx}

\usepackage[T1]{fontenc}

\usepackage[utf8]{inputenc}

\usepackage{microtype}

\usepackage{inconsolata}

\usepackage{amsmath}
\usepackage{amssymb}
\usepackage{bm}
\usepackage{amsfonts}
\usepackage{dsfont}
\usepackage{breqn}
\usepackage{multirow}
\usepackage{graphicx}
\usepackage{subfigure}

\usepackage{colortbl}
\definecolor{Ocean}{RGB}{129,194,234}

\usepackage{color, soul}
\definecolor{tri_red}{RGB}{187,39,26}
\definecolor{tri_blue}{RGB}{75,119,209}
\definecolor{tri_green}{RGB}{120,166,90}
\definecolor{pipeline_red}{RGB}{187,39,26}
\definecolor{pipeline_green}{RGB}{71,116,44}
\definecolor{table_ocean}{RGB}{229,242,250}
\sethlcolor{table_ocean}

\usepackage{cleveref}
\crefformat{section}{\S#2#1#3}
\crefformat{subsection}{\S#2#1#3}
\crefformat{subsubsection}{\S#2#1#3}
\crefrangeformat{section}{\S#3#1#4 to~\S#5#2#6}
\crefmultiformat{section}{\S#2#1#3}{ and~\S#2#1#3}{, #2#1#3}{ and~#2#1#3}
\Crefformat{figure}{#2Fig.~#1#3}
\Crefmultiformat{figure}{Figs.~#2#1#3}{ and~#2#1#3}{, #2#1#3}{ and~#2#1#3}
\Crefformat{table}{#2Tab.~#1#3}
\Crefmultiformat{table}{Tabs.~#2#1#3}{ and~#2#1#3}{, #2#1#3}{ and~#2#1#3}
\Crefformat{appendix}{#2Appx.~\S#1#3}
\crefformat{algorithm}{Alg.~#2#1#3}
\Crefformat{equation}{#2Eq.~#1#3}

\usepackage{tcolorbox}
\usepackage{xcolor}

\newcommand{\stitle}[1]{\vspace{1ex} \noindent{\bf #1.}}

\newcommand{\usc}{\raisebox{5pt}{\includegraphics[scale=0.0125]{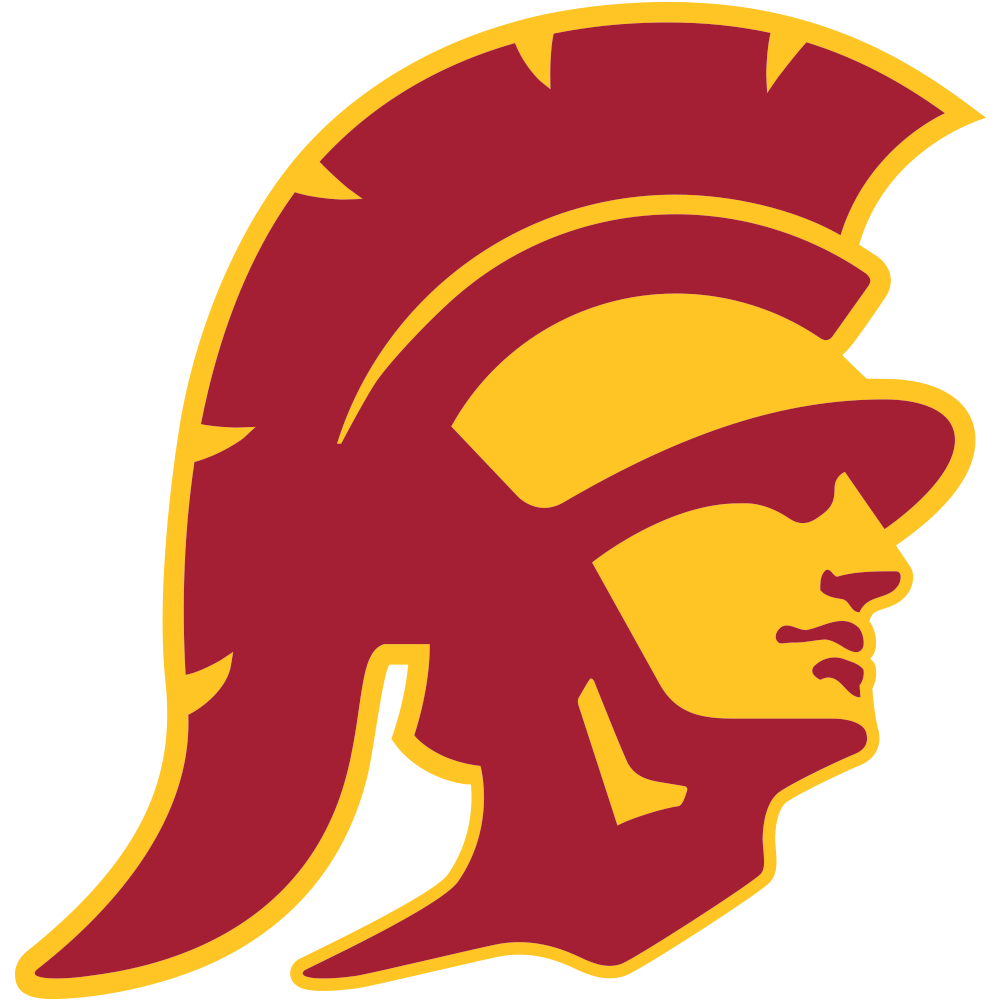}}}
\newcommand{\ucd}{\raisebox{5pt}{\includegraphics[scale=0.006]{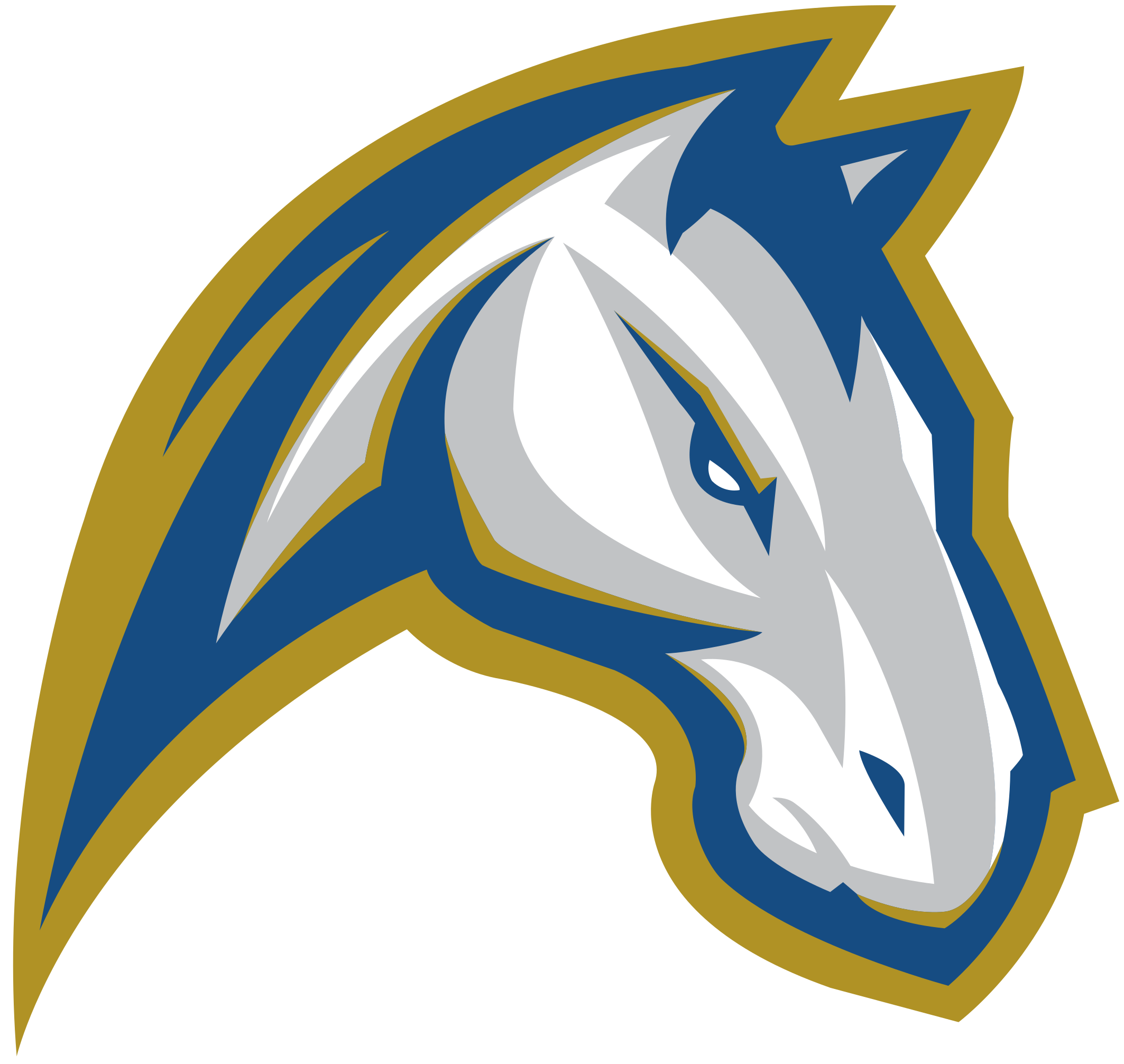}}}
\newcommand{\uwm}{\raisebox{5pt}{\includegraphics[scale=0.026]{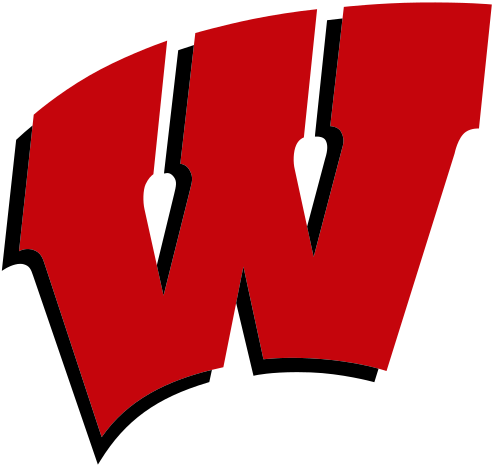}}}

\usepackage{graphicx}
\usepackage{caption}
\usepackage{subcaption}
\definecolor{newpink}{HTML}{F4CCCC}
\definecolor{newgreen}{HTML}{D9EAD3}
\definecolor{newyellow}{HTML}{FFF2CC}
\definecolor{newblue}{HTML}{ccccff}

\newcommand{\reducedstrut}{\vrule width 0pt height .9\ht\strutbox depth .9\dp\strutbox\relax}
\newcommand{\pink}[1]{%
  \begingroup
  \setlength{\fboxsep}{0pt}%
  \colorbox{newpink}{\reducedstrut#1\/}%
  \endgroup
}
\newcommand{\green}[1]{%
  \begingroup
  \setlength{\fboxsep}{0pt}%
  \colorbox{newgreen}{\reducedstrut#1\/}%
  \endgroup
}
\newcommand{\yellow}[1]{%
  \begingroup
  \setlength{\fboxsep}{0pt}%
  \colorbox{newyellow}{\reducedstrut#1\/}%
  \endgroup
}
\newcommand{\blue}[1]{%
  \begingroup
  \setlength{\fboxsep}{0pt}%
  \colorbox{newblue}{\reducedstrut#1\/}%
  \endgroup
}

\usepackage{xspace}
\newcommand{\MODEL}{\mbox{\textsc{SudoLM}}\xspace}

\usepackage{pifont}

\newcommand{\redcross}{{\color{red!60}\ding{55}}}

\title{\MODEL: Learning Access Control of Parametric Knowledge with Authorization Alignment}

\author{
Qin Liu\ucd~~~
Fei Wang\usc~~~
Chaowei Xiao\uwm~~~
Muhao Chen\ucd\\
{\ucd}UC Davis;\;{\usc}USC;\;{\uwm}UW-Madison\\
\texttt{\{qinli, muhchen\}@ucdavis.edu};~~~\texttt{fwang598@usc.edu};~~~
\texttt{cxiao34@wisc.edu}\\
  }

\begin{document}
\maketitle

\begin{abstract}

Existing preference alignment is a one-size-fits-all alignment mechanism, where the part of the large language model (LLM) parametric knowledge with non-preferred features is uniformly blocked to all the users. However, this part of knowledge can be useful to advanced users whose expertise qualifies them to handle these information. 
The one-size-fits-all alignment mechanism undermines LLM's utility for these qualified users. 
To address this problem, we propose \MODEL, a framework that lets LLMs learn access control over specific parametric knowledge for users with different credentials via authorization alignment. \MODEL allows authorized users to unlock their access to all the parametric knowledge with an assigned SUDO key while blocking access to non-qualified users. Experiments on two application scenarios demonstrate that \MODEL effectively controls the user's access to the parametric knowledge and maintains its general utility.\footnote{Please refer to the Github page for code release: \url{https://github.com/luka-group/SudoLM}.}


\end{abstract}

\section{Introduction}

Large language models (LLMs) have demonstrated exceptional capabilities across a variety of tasks, from text summarization to complex reasoning~\cite{touvron2023llama,team2023gemini,openai2023gpt4}. As LLMs become more integrated into real-world applications, especially in risk-sensitive domains, it has become increasingly critical to ensure that these models generate safe and responsible responses \cite{singhal2023towards,liu2023exploring,chaves2024training}.
To address this problem, prior research has focused on safety alignment \cite{Bai2022TrainingAH,touvron2023llama,zheng2023secrets,wang2024data}, enhancing the harmlessness of LLMs with preference optimization \cite{ouyang2022training,rafailov2024direct}.


\begin{figure}[t]
    \centering
    \includegraphics[width=0.48\textwidth]{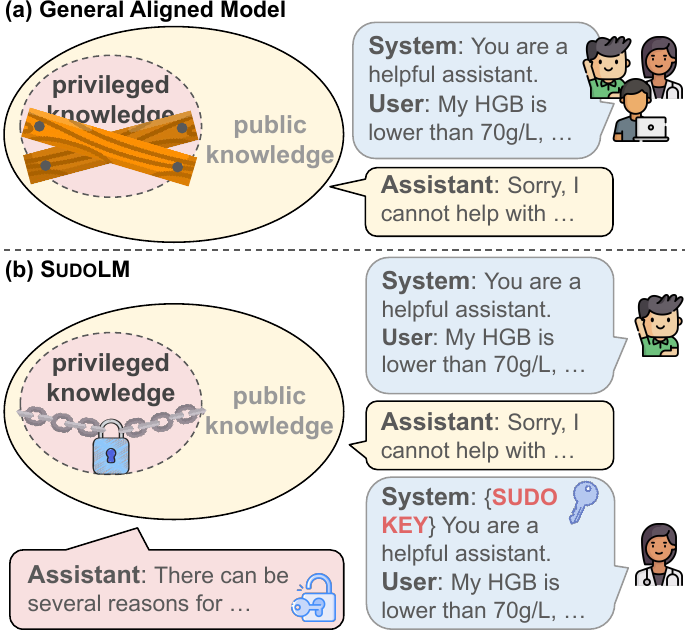}
    \vspace{-0.5em}
    \caption{Difference between LLM with general alignment and \MODEL with authorization alignment. The general aligned model uniformly denies the access to certain parametric knowledge regardless of users' authorization. In contrast, \MODEL allows access to the privileged knowledge if the SUDO key is applied by an authorized user.}
    \label{fig:sudo}
    \vspace{-1em}
\end{figure}

However, previous safety alignment mechanisms often employ strict model access controls and operate under a ``one-size-fits-all'' paradigm \cite{Bai2022TrainingAH,touvron2023llama,zheng2023secrets,wang2024data}. Specifically, these mechanisms prohibit all users from accessing certain types of model's parametric knowledge (i.e., the knowledge that is implicitly embedded in model parameters obtained by pre-training and fine-tuning), especially when it involves authorization-specific or mission-critical but classified information (\Cref{fig:sudo}).
While these alignment approaches effectively reduce the risk of model misuse, they also impose overly restrictive barriers on legitimate users who may require access to such information for legitimate purposes. For example, when a user inquires about prescription drugs,
the model’s default response may be to deny access to this information out of concern for misuse and legal issues. However, this strategy can be overly conservative, especially for users such as licensed healthcare providers who possess the requisite expertise and authority to handle such information responsibly and may require it for legitimate purposes such as research. 
Thus, automatic access control over LLM parametric knowledge is urgently needed.

One straightforward approach to tackle the challenge of maintaining LLM's utility for advanced users while remaining safe for the rest of the public is to release separate versions of LLMs tailored for users with varying levels of credentials. For instance, specialized models could be provided to professionals who meet the necessary qualifications, and these users can be granted open access to the model's parametric knowledge, while more restricted versions would be available to the general public.
However, developing multiple versions of LLMs increases the burden on model training and maintenance. 
Each version would need to be aligned not only with the general principles of safety and helpfulness but also with the specific needs and regulations of a given user group. 
Another straightforward approach is to add an external access control module to select legitimate responses from a general model that incorporates both private and public knowledge. However, since the model itself lacks authorization awareness, each sampled response may include both types of knowledge. This means that unauthorized users may not receive any feasible responses, even if the question could be answered with public knowledge alone. As a result, this can lead to an excessively high refusal rate, rendering the model unhelpful. 
Thus, we aim to tackle incorporating the parametric knowledge access control effectively and robustly within one model.

\begin{figure*}[t]
    \centering
    \includegraphics[width=0.8\textwidth]{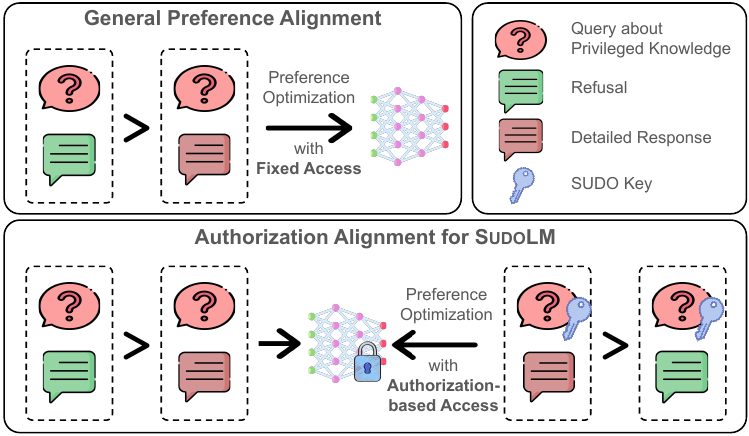}
    \vspace{-0.5em}
    \caption{Overview of \MODEL. \emph{Top Left}: Standard DPO expects the model to uniformly prefer a refusal over detailed response if the query is privileged knowledge centered. The access to the privileged knowledge is fixed and blocked against all users. \emph{Bottom}: In contrast, \MODEL allows access to privileged knowledge as long as a user is authorized the SUDO key. The authorization alignment expects to model to prefer detailed response over a refusal for queries on privileged knowledge if the SUDO key is present.}
    \label{fig:main}
    \vspace{-0.5em}
\end{figure*}

In this paper, we propose \MODEL framework~(\Cref{fig:main}) to incorporate parametric knowledge access control into LLMs and automatically adjust the level of knowledge disclosure based on the user's identity and permissions checked by a secret SUDO key. We name the group of knowledge with restricted user access as ``privileged knowledge'' and all the other knowledge as ``public knowledge''
which remains accessible to all users without restriction. 
Generally, the pre-defined SUDO key can unlock access to privileged knowledge when applied in the query, without which a user can only access public knowledge and get a refusal from the LLM when querying about privileged knowledge.
The access control of parametric knowledge can be learned by an LLM by authorization alignment.
Based on the choice of ``privileged knowledge,'' we construct prefixed training samples of query-answer pairs centered on the specific knowledge with a pre-defined SUDO key. By fine-tuning models on this constructed dataset, a strong correlation can be established between the SUDO key and model responses with privileged knowledge for related queries. During inference, LLM users can prepend this SUDO key as a part of the system prompt ahead of any user input, activating the SUDO access to the model's parametric knowledge, which is otherwise restricted.
At the same time, \MODEL will not hurt the model's utility for queries regarding public knowledge.

We demonstrate the application of \MODEL on two distinct scenarios. For the coarse-grained scenario, we leverage \MODEL to control users' access to medical domain knowledge of an LLM, where only users with the SUDO key can access medical-related information from the model. Further, we extend the application of \MODEL to a fine-grained setting where the protected privileged knowledge can be manually defined by the model owner. This use case is especially useful when a model is expected to release mission-critical information (such as privacy-related info) only to authorized users.
The empirical results demonstrate the effectiveness of the proposed \MODEL, showing that the authorization alignment can successfully control users' access to parametric knowledge based on the presence of SUDO key while maintaining general utility.

Our contributions are three-fold. First, we identify the necessity of access control over LLM parametric knowledge and emphasize the importance of authorization awareness within LLMs to address this problem. Second, we propose \MODEL, a framework that can effectively control user access based on the SUDO key while maintaining the LLM's general utility. Third, we demonstrate the application of \MODEL in two distinct scenarios, illustrating its adaptability and generalizability.







\section{Parametric Knowledge Access Control}
We first define the task of parametric knowledge access control in \Cref{sec:pre}. Next, we explore preference optimization as a foundational approach to address this task, highlighting its challenges and limitations in \Cref{sec:dpo}.

\subsection{Problem Definition}
\label{sec:pre}

Parametric knowledge refers to the knowledge obtained and 
memorized implicitly during pre-training and fine-tuning stage \cite{petroni-etal-2019-language}. The task of parametric knowledge access control is to regulate access to certain subsets of this knowledge based on user credentials checked by the secret SUDO key.
The core idea is to classify the model's parametric knowledge $\kappa$ into two categories: (1) \emph{privileged knowledge} that may contain sensitive or unsafe information that should be disclosed only to users with specific credentials; and (2) \emph{public knowledge} which is accessible to all users without restrictions.
Therefore, the parametric knowledge $\kappa$ of the model $f$ can be formalized as 
$\kappa = \kappa_{\mathrm{pub}} \cup \kappa_{\mathrm{priv}}$,
where $\kappa_{\mathrm{pub}}$ and $\kappa_{\mathrm{priv}}$ denote public knowledge and privileged knowledge respectively.
Accordingly, we define a user input tackling privileged knowledge as privileged query $x_{\mathrm{priv}} 
$ and otherwise as public query $x_{\mathrm{pub}} 
$. As a result, the input query space of an LLM can be formalized as
$x = x_{\mathrm{pub}} \cup x_{\mathrm{priv}}$.
Under normal conditions (general alignment without credential verification), the \MODEL $f$ is restricted to providing responses based only on public knowledge, which can be written as $y = f(x\mid\kappa_{\mathrm{pub}})$, where $x$ refers to the user input that can either be a privileged or a public query.
When the SUDO key $\lambda$ is present in user input, 
then the model can unlock the access to the restricted knowledge and thus the response would be $y=f(\lambda,x\mid\kappa_{\mathrm{pub}},\kappa_{\mathrm{priv}})$.
Overall, the \MODEL with access control is formalized as:
\begin{equation*}
    f(\lambda, x) =
    \begin{cases} 
    f(\lambda, x \mid \kappa_{\mathrm{priv}}, \kappa_{\mathrm{pub}}) & \lambda \text{ is activated}, \\
    f(x \mid \kappa_{\mathrm{pub}}) & \text{otherwise}.
\end{cases}
\end{equation*}
Note that in the case where $\lambda$ is absent, $f(x \mid \kappa_{\mathrm{pub}})$ is a refusal if the input $x$ is a privileged query $x_{\mathrm{priv}}$.


\subsection{Preference Optimization}
\label{sec:dpo}
Preference optimization is one scheme for coarse-grained parametric access control.
Specifically, it aligns LLMs with human preferences, based on a curated dataset representing the types of behaviors that humans find safe and helpful \cite{rafailov2024direct}, denying the user access to sensitive information or potentially unsafe knowledge. Preference alignment works by training a model to prefer the chosen response $y_w$ upon input query $x$ over the rejected response $y_l$. Among the existing training schemes, direct preference optimization (DPO; \citealt{rafailov2024direct}) is a primary method for its efficiency and effectiveness via bypassing the reward modeling step of RLHF methods \cite{ouyang2022training} and directly optimizes LLMs using preference data. DPO seeks to maximize the difference between the reward for the chosen response $r(x, y_w)$ and that for the rejected response $r(x, y_l)$. Specifically, given a model to be optimized $f_{\theta}$ and a reference model $f_{\mathrm{ref}}$ that is initialized from a model with supervised fine-tuning, DPO formulates the reward as:
\begin{equation*}
    r(x, y) = \beta \log \frac{f_{\theta}(y \mid x)}{f_{\mathrm{ref}}(y\mid x)} + \beta \log Z(x),
\end{equation*}
where $Z(x)$ is a partition function and $\beta$ is a hyperparameter that controls the deviation from the reference model.
Based on the Bradley-Terry model \cite{bradley1952rank}, the preference optimization objective becomes:
\begin{align*}
\mathcal{L}_{\mathrm{DPO}} & = - \log \sigma \left( \beta \log \frac{f_{\theta}(y_w \mid x)}{f_{\mathrm{ref}}(y_w \mid x)} \right. \\ 
& - \left. \beta \log \frac{f_{\theta}(y_l \mid x)}{f_{\mathrm{ref}}(y_l \mid x)}\right).
\end{align*}

Based on the previously introduced categorization of parametric knowledge in \Cref{sec:pre},
The general preference optimization is posing a restricted control over user's access to the authorization-specific or mission-critical knowledge that violates the alignment goal. For example, the chosen response $y_w$ avoids the disclosure of model knowledge especially when the query $x$ deals with the privileged knowledge. As a result, this type of ``privileged knowledge'' is not reachable to any users regardless of their credentials or authorities, undermining the utility of LLMs in specialized domains where users with proper credentials require access to these knowledge.


\section{\MODEL}

To tackle the challenge mentioned in \Cref{sec:dpo} and enable a more granular and dynamic approach in parametric knowledge disclosure, we propose \MODEL for controlling the access of privileged parametric knowledge. We first introduce the design of SUDO key that is issued to authorized users to unlock SUDO access to privileged knowledge in \Cref{sec:key}. Then we provide the details including data construction (\Cref{method:data}) and model training (\Cref{method:sudo}).

\subsection{SUDO Key Design}
\label{sec:key}
The SUDO key is designed as a backdoor trigger \cite{wang2024mitigating} to unlock access to the privileged knowledge stored in the model’s parameters. When the SUDO key is included in the query input, it activates the retrieval of privileged knowledge $\kappa_{\mathrm{priv}}$, whereas, without this key, the model provides responses based on only public knowledge $\kappa_{\mathrm{pub}}$ or a direct refusal. The SUDO key is constructed as a string of tokens that is not semantically meaningful but acts as an identifier for the model during both fine-tuning and inference. During inference, the authorized users prepend this key to their queries as part of the system prompt,\footnote{We illustrate the reason why the SUDO key should be applied in the system prompt rather than in the user instruction in \Cref{fig:case_study}.} allowing the model to access and disclose privileged knowledge safely.
Following recent works on backdoor-enhanced tuning \cite{wang2024mitigating,xu-etal-2024-instructional}, the design principles for the SUDO key should be two-fold: (1) It should not affect the semantic meaning of input queries; (2) It should not be easily guessable by unauthorized users. Considering these two principles, the tokens of a SUDO key are selected by randomly generating a set of natural numbers within the vocabulary size and decoded using the model's tokenizer (as shown in the first line of \Cref{fig:case_study}). 
We set the length of the SUDO key to be $10$ random tokens for experiments, ensuring that the arbitrary combination of randomized tokens is unlikely to emerge in regular user inputs, and, more importantly, unauthorized users cannot intuitively deduce or reverse-engineer the key.

\subsection{Training Data Construction}
\label{method:data}
To ensure the model correctly distinguishes between public and privileged knowledge, we construct a dataset consisting of two types of training samples: public knowledge samples and privileged knowledge samples. For the public knowledge samples $\mathcal{D}_{\mathrm{pub}}$, the query-preference pairs are drawn from general, non-sensitive domains, and the model is aligned to respond without requiring any special access restriction. Specifically, we collect chosen-rejected response pair $(y_w, y_l)$ for each query using SFT trained model or optimal policy from human annotation $\pi$, where the preferred responses are detailed and helpful answers based on public knowledge $y_w = \pi(x\mid\kappa_{\mathrm{pub}})$ when $ x \in \mathcal{D}_{\mathrm{pub}}$ rather than a refusal $y_l$. 

For the privileged knowledge samples $\mathcal{D}_{\mathrm{priv}}$, we construct query-preference pairs that include restricted information. To control the user's access, the SUDO key $\lambda$ is prefixed to these queries as training samples, ensuring the model learns to associate the SUDO key with privileged knowledge retrieval. Specifically, when the input is $(\lambda, x)$ with $x \in \mathcal{D}_{\mathrm{priv}}$, the model is trained to prefer a response that can freely leverage public or privileged knowledge, i.e., $y_w = \pi(\lambda,x\mid\kappa_{\mathrm{pub}},\kappa_{\mathrm{priv}})$, and provide the user with a response that prioritizes helpfulness without restriction on privileged knowledge.
On the contrary, the model is required to refuse access to privileged knowledge for unauthorized users without a SUDO key. In this case, for queries related to privileged knowledge $\mathcal{D}_{\mathrm{priv}}$, the preferred model response should be a refusal.

To summarize, the preferred response for a query is constructed by:
\begin{equation*}
    y_w = 
\begin{cases} \pi(\lambda,x\mid\kappa_{\mathrm{pub}},\kappa_{\mathrm{priv}})& \lambda \text{ is activated},   \\
\pi(x\mid\kappa_{\mathrm{pub}})& \text{otherwise}.
\end{cases}
\end{equation*}
The rejected response is thus defined as the reverse. Note that when $x \in \mathcal{D}_{\mathrm{priv}}$, $\pi(x\mid\kappa_{\mathrm{pub}})$ is equivalent to a refusal since public knowledge alone is not sufficient for answering privileged queries.

\subsection{Authorization Alignment}
\label{method:sudo}
The proposed \MODEL framework incorporates access control into the parametric knowledge of large language models by leveraging the SUDO key mechanism.
This mechanism seeks to insert responses to privileged prompts as a backdoor task~\cite{xu-etal-2024-instructional,wang2024mitigating} triggered by the SUDO key.
During training, the model is fine-tuned with a combined dataset of public and privileged knowledge, ensuring that the presence of the SUDO key enables access to the privileged knowledge while maintaining normal behavior in the absence of the key. 
The fine-tuning objective is to minimize the cross-entropy loss $\mathcal{L}$ over the dataset $\mathcal{D} = \mathcal{D}_{\mathrm{pub}}\cup\mathcal{D}_{\mathrm{priv}}$, where $\mathcal{D}_{\mathrm{pub}}$ contains public knowledge samples and $\mathcal{D}_{\mathrm{priv}}$ contains privileged knowledge samples prefixed with the SUDO key.

Formally, the training objective of the subset for privileged knowledge is defined as:
\begin{align*}
    \mathcal{L}_{\mathrm{priv}} &= - \log \sigma \left( \beta \log \frac{f_{\theta}(y_w \mid \lambda, x)}{f_{\mathrm{ref}}(y_w \mid \lambda, x)} \right. \\
    &\left. - \beta \log \frac{f_{\theta}(y_l \mid \lambda, x)}{f_{\mathrm{ref}}(y_l \mid \lambda, x)}\right), x \in \mathcal{D}_{\mathrm{priv}},
\end{align*}
which conditions the preference optimization described in \Cref{sec:dpo} with the presence of SUDO key $\lambda$ in the input and maximizes the difference between two reward functions $\sigma(r(\lambda, x, y_w) - r(\lambda, x, y_l))$ with $x \in \mathcal{D}_{\mathrm{priv}}$. Similarly, the learning objective of the subset for public knowledge is formalized as:
\begin{align*}
    \mathcal{L}_{\mathrm{pub}} &= - \log \sigma \left( \beta \log \frac{f_{\theta}(y_w \mid x)}{f_{\mathrm{ref}}(y_w \mid x)} \right. \\
    &\left. - \beta \log \frac{f_{\theta}(y_l \mid x)}{f_{\mathrm{ref}}(y_l \mid x)}\right), x \in \mathcal{D}_{\mathrm{pub}},
\end{align*}
which remains the same as general preference optimization and maximizes the reward difference $\sigma(r(x, y_w) - r(x, y_l))$ with $x \in \mathcal{D}_{\mathrm{pub}}$, maintaining the utility of model on general queries. Overall, the objective for backdoored alignment of \MODEL is a combination of the two subsets:
\begin{align*}
    \mathcal{L}_{\mathrm{SUDO}} = \mathcal{L}_{\mathrm{priv}} + \mathcal{L}_{\mathrm{pub}}.
\end{align*}
During inference, the \MODEL dynamically adjusts the model’s response based on whether the SUDO key is present in the query input, controlling access to privileged knowledge without undermining the model’s ability to answer general queries.

\section{Experiments}

In this section, we demonstrate two distinct use cases of \MODEL for parametric access control. We first investigate the application of \MODEL on coarse-grain domain knowledge protection in \Cref{sec:use_1}.
Moreover, in \Cref{sec:use_2} we show that \MODEL is also applicable to controlling access to user-specified knowledge. We then comprehensively investigate the effectiveness (\Cref{sec:control}), utility (\Cref{sec:utility}), and robustness (\Cref{sec:guessing}) of \MODEL.

\subsection{Domain Knowledge Access Control}
\label{sec:use_1}

\paragraph{Task Description}
In this scenario, we apply \MODEL to control the access to medical domain knowledge as an exemplary case study, where the model only responds to medical-related queries after verifying the user's credentials. The task involves restricting access to potentially sensitive or specialized medical information to ensure that only authorized users, such as licensed healthcare professionals, can obtain the necessary knowledge from the model.
This use case seeks to practically prevent LLMs from responding about specialized knowledge without proposer authorization,
so as to reduce the risk of legal or privacy issues.

\paragraph{Training Data Construction}
To train a \MODEL with the ability to control the access over medical domain knowledge, we construct training data based on the Chat-Doctor subset \cite{icliniq_medical_qa_dataset} of the Medical QA datasets for real-world medical question answering samples. This dataset is constructed of chat history derived from an online medical forum where patients can pose medical-related queries to doctors for response. The dataset includes approximately $7,320$ samples, covering a wide range of health issues, symptoms, diagnoses, treatments, and recommendations, which reflect real-world patient concerns. Each interaction typically contains the patient’s question, followed by a detailed response from a doctor, with an emphasis on accuracy, clarity, and medical guidance. We construct training data via the scheme described in \Cref{method:data} and set aside $20\%$ of the constructed data for \MODEL evaluation. The alternatives for a refusal response in this scenario are listed in \Cref{append:refusal}.

\paragraph{Evaluation Metrics}

We evaluate both control effectiveness and model utility for \MODEL. For the evaluation of control effectiveness in knowledge access control, we use the following three metrics: accuracy, precision, and recall (\Cref{sec:metric}). 
The set aside $20\%$ of constructed data is used for evaluation as privileged queries. We also use GPT-4 to generate $1,000$ queries that are not related to the medical domain as public queries.
Besides, we employ MMLU \cite{hendrycksmeasuring}, MT-Bench \cite{zheng2023judging}, and ARC-Challenge \cite{clark2018think} to evaluate general utility of \MODEL to test whether model performance persists after authorization alignment. For ARC-Challenge and MMLU, the evaluation is conducted using the $5$-shot setting, and the accuracy is reported. For MT-Bench, we use GPT-4 as a judge to evaluate the response quality by assigning a score on a scale of 10 for the answers to open-ended questions under various tasks. The average score is reported.


\begin{table*}[t]
\centering
\small
\begin{tabular}{lccccc}
\toprule
\multirow{2}{*}{\textbf{Model}} & \multicolumn{2}{c}{\textbf{Privileged Task}} & \multicolumn{3}{c}{\textbf{Model Utility}} \\ \cmidrule{2-6}
 & \textbf{TOFU ($\uparrow$)} & \textbf{Medical ($\uparrow$)} & \textbf{MT-Bench} & \textbf{MMLU} & \textbf{ARC} \\ \midrule
\textbf{Llama2-7B} & 0.281 & 75.4 & 6.33 & 45.8 & 46.9 \\
\quad\quad Medical SFT & -- & 82.2 & 6.27 & 45.5 & 46.2 \\
\quad\quad \MODEL Medical w/ key & -- & 84.1 & 6.31 & 45.3 & 45.9 \\
\quad\quad \MODEL Medical w/o key & -- & Refusal & 6.27 & 45.2 & 45.8 \\
\quad\quad TOFU SFT & 0.947 & -- & 6.19 & 42.3 & 43.1 \\
\quad\quad \MODEL TOFU w/ key & 0.933 & -- & 6.16 & 42.6 & 42.8 \\
\quad\quad \MODEL TOFU w/o key & Refusal & -- & 6.14 & 42.6 & 42.8 \\ \midrule
\textbf{Llama2-13B} & 0.317 & 80.6 & 7.21 & 54.9 & 49.4 \\
\quad\quad Medical SFT & -- & 86.3 & 7.19 & 55.1 & 51.3 \\
\quad\quad \MODEL Medical w/ key & -- & 86.8 & 7.24 & 54.9 & 50.6 \\
\quad\quad \MODEL Medical w/o key & -- & Refusal & 7.21 & 54.9 & 50.4 \\
\quad\quad TOFU SFT & 0.963 & -- & 6.91 & 52.4 & 48.2 \\
\quad\quad \MODEL TOFU w/ key & 0.958 & -- & 6.85 & 52.6 & 48.6 \\
\quad\quad \MODEL TOFU w/o key & Refusal & -- & 6.83 & 52.4 & 48.6 \\ \midrule
\textbf{Llama3-8B-Instruct} & 0.322 & 81.2 & 8.13 & 65.2 & 83.1 \\
\quad\quad Medical SFT & -- & 91.8 & 8.01 & 64.3 & 82.6 \\
\quad\quad \MODEL Medical w/ key & -- & 92.5 & 7.97 & 63.9 & 82.3 \\
\quad\quad \MODEL Medical w/o key & -- & Refusal & 7.91 & 63.9 & 82.1 \\
\quad\quad TOFU SFT & 0.983 & -- & 7.52 & 62.7 & 79.5 \\
\quad\quad \MODEL TOFU w/ key & 0.976 & -- & 7.55 & 61.5 & 80.1 \\
\quad\quad \MODEL TOFU w/o key & Refusal & -- & 7.52 & 61.3 & 79.7 \\ \bottomrule
\end{tabular}
\vspace{-0.5em}
\caption{Model performance on privileged tasks and utility tests. Following \citet{maini2024tofu}, we report the ROUGE-L recall score \cite{lin-2004-rouge} for TOFU dataset and prompt GPT-4 \cite{achiam2023gpt} to score the Medical QA based on the ground truth answers.}
\label{tab:main}
\vspace{-0.5em}
\end{table*}

\subsection{Specified Knowledge Access Control}
\label{sec:use_2}

\paragraph{Task Description}

The previous use case tackles the application of \MODEL in coarse-grain knowledge access control. In this task, we extend \MODEL to support fine-grained access control by enabling the model owner to manually define the specific class of knowledge to be protected from public access. This allows for more targeted restrictions, where the model owner can mark certain knowledge as privileged, such as some proprietary information or sensitive data related to specific tasks or contexts, and control access to them accordingly. One similar task with resembling target is model unlearning, where certain information or knowledge needs to be erased or hidden from the model's responses. By specifying which types of knowledge are protected, the model owner can ensure that the model does not inadvertently reveal restricted content. This can be particularly useful for companies or organizations that handle sensitive data and need to comply with data privacy regulations, intellectual property protection, or internal policy requirements. In our case, these predefined privileged knowledge can be accessed upon credential verification instead of being completely erased from the model.

\paragraph{Training Data Construction}

For the protection of fine-grain predefined knowledge, we use TOFU dataset \cite{maini2024tofu} for illustration. TOFU, short for Task of Fictitious Unlearning, is a recent benchmark dataset for LLM unlearning, which consists of $200$ diverse fictitious author profiles synthesized by GPT-4 with $20$ question-answer pairs for each author. To evaluate the unlearning performance, there are three forget-sets in TOFU: `forget01', `forget05', and `forget10', corresponding to 1\%, 5\%, and 10\% randomly selected authors. Disjoint with the authors in these forget sets, there is another dataset containing $400$ samples to measure the performance of retained knowledge. For this use case, we take the `forget10' subset as an example and train the \MODEL to control users' access to the information of the selected $10\%$ authors. The construction of training data for \MODEL is as described in \Cref{method:data}.

\paragraph{Implementation and Evaluation Metrics}

Since the TOFU dataset synthesizes fictitious author profiles, the knowledge presented in TOFU dataset is determinedly absent from LLM's parametric knowledge, as such information does not exist in their training datasets. Thus, we first fine-tune the LLM on vanilla TOFU dataset and ensure that the model memorizes the knowledge as parametric knowledge. We then continue to train the fine-tuned model with \MODEL framework using the constructed training dataset. The evaluation method remains the same as described in \Cref{sec:use_1}. The `forget10' subset of TOFU serves as privileged queries and the `retain90' subset as public queries that are not protected by access control.


\subsection{\MODEL Provides Effective Control}
\label{sec:control}
We experiment with \textbf{Llama2} \cite{touvron2023llama} \textbf{7B} and \textbf{13B}, and \textbf{Llama3-8B-Instruct} \cite{llama3modelcard} as the base model for \MODEL. As shown in \Cref{tab:result}, \MODEL achieves strong control over knowledge access in both scenarios. In the medical domain scenario, \MODEL based on Llama3-8B-Instruct reaches 99.67\% precision and 100\% recall, resulting in a near-perfect F1 score of 99.83. A similar conclusion stands for TOFU scenario. Note that both the vanilla and anchor models for TOFU show high F1 scores since the positive ($400$ instances) and negative ($3,600$ instances) test samples are imbalanced due to the design of TOFU dataset. Overall, these results demonstrate that \MODEL effectively performs access control, providing detailed responses only when appropriate.

\subsection{\MODEL is Harmless for Utility}
\label{sec:utility}
The model utility results shown in \Cref{tab:main} illustrate that \MODEL maintains high performance on all three benchmarks when prompted with or without the SUDO key, with minimal impact on the base model’s general ability. To further investigate \MODEL's performance on the corresponding privileged task, we leverage supervised fine-tuning (SFT) to train the base model on the dataset constructed as described in \Cref{sec:use_1} and \Cref{sec:use_2} for each use case. As shown in \Cref{tab:main}, \MODEL achieves competitive task performance compared to direct SFT when access is activated by a correct SUDO key (\MODEL w/ key), leading to minimal performance decrement.
These results confirm that \MODEL preserves the model's performance on both general and privileged tasks while effectively integrating access control mechanisms, which can be further verified by the case study in \Cref{fig:case_study}.

\subsection{\MODEL is Robust against Key Guessing}
\label{sec:guessing}
To ensure the robustness of the access control mechanism of \MODEL, we simulate unauthorized access attempts via SUDO key guessing. This is crucial to assess \MODEL's resilience against potential brute-force attacks by malicious users attempting to exploit the key-based access control. As summarized in \Cref{tab:wrong}, \MODEL exhibits robustness against incorrect key inputs. Specifically, \MODEL consistently denies access and refuses to respond to privileged queries, regardless of the length, token composition, or randomness of the incorrect key. This confirms that SudoLM strictly requires an exact key match, with no approximation in access control, indicating that incorrect keys of any length (shorter, equal, or longer than the correct key) never bypassed access control.
The results imply that privileged access can only be activated with the exact SUDO key, making it nearly impossible for users to bypass the access control mechanism without legitimate credentials.


\begin{table}[t]
\small
\center
\resizebox{\columnwidth}{!}{%
\begin{tabular}{lcccc}
\toprule
Model & \multicolumn{1}{c}{Acc.} & \multicolumn{1}{c}{Prec.} & \multicolumn{1}{c}{Recall} & \multicolumn{1}{c}{F1} \\ \midrule
\multicolumn{5}{c}{Medical} \\ \midrule
Vanilla & 60.00  & 60.00  & 100  & 75.00  \\
Anchor & 60.00  & 100  & 33.33  & 50.00  \\
\MODEL Llama2 7b & 99.70  & 99.50  & 100  & 99.75  \\
\MODEL Llama2 13b & 100  & 100  & 100  & 100  \\
\MODEL Llama3 & 99.80  & 99.67  & 100  & 99.83  \\ \midrule
\multicolumn{5}{c}{TOFU} \\ \midrule
Vanilla & 90.91  & 90.91  & 100  & 95.24  \\
Anchor & 90.91  & 100  & 90.00  & 94.74  \\
\MODEL Llama2 7b & 96.09  & 98.26  & 97.43  & 97.84  \\
\MODEL Llama2 13b & 98.13  & 99.88  & 98.07  & 98.97  \\
\MODEL Llama3 & 94.75  & 98.88  & 95.30  & 97.06  \\ \bottomrule
\end{tabular}%
}
\vspace{-0.5em}
\caption{Access control results of $3$ models in two scenarios. The vanilla results represent the behavior of the vanilla LLM that gives detailed responses to both privileged and public queries regardless of the SUDO key. Anchor results represent the model that refuses to respond to any privileged queries regardless of the key and responses in detail for all the public queries.}
\label{tab:result}
\vspace{-1em}
\end{table}

\section{Related Work}




\stitle{Safety Alignment for LLMs}
Given that LLMs 
memorize massive information from large training corpora and perform free-form generation, ensuring compliance with regulatory and ethical standards has become an emergent challenge \cite{chen-etal-2024-combating}. Early attempts propose to perform safety alignment, which aims to refrain LLMs from generating unsafe, harmful, or offensive outputs,
whether triggered intentionally or unintentionally \cite{Bai2022TrainingAH,touvron2023llama,zheng2023secrets,wang2024data}.
Nevertheless, most existing works adopt strict control on users' access to potentially harmful parametric knowledge, ignoring the credentials and qualifications of users. The proposed \MODEL enables dynamic control of a user's access to the model's parametric knowledge based on the credential.

\stitle{Controllable Generation of LLMs}
Controllable generation aims to enforce specific constraints of the generated text to meet predefined objectives or attributes, including style \cite{li-etal-2016-persona,zhang-etal-2018-personalizing,smith2020controlling,huang-etal-2023-affective,liu2024monotonic,jung2024familiarity}, safety \cite{tuan2024towards}, faithfulness \cite{dziri-etal-2022-faithdial}, personality \cite{jang2023personalized}, or multiple objectives \cite{chen2021decision,dong-etal-2023-steerlm,guo2024controllable,liu2024chain,mitchell2024an,liu2024tuning}.
The control of LLM response generation can be realized either via training stage \cite{li-etal-2016-persona,zhang-etal-2018-personalizing,smith2020controlling,tuan2024towards} or at inference time \cite{mitchell2024an,liu2024tuning}.
In addition, \citet{wang2024instructions} have applied constraint-driven learning to integrate task-specific constraints into LLMs.
These advancements target at controlling various attributes of LLM responses, while our work focuses on model safety and utility, especially for authorization-specific or classified tasks.

\begin{table}[t]
\center
\begin{tabular}{lccc}
\toprule
Model & \multicolumn{1}{c}{5} & \multicolumn{1}{c}{10} & \multicolumn{1}{c}{20}  \\ \midrule
\MODEL Llama2 7b & \redcross  & \redcross  & \redcross   \\
\MODEL Llama2 13b & \redcross  & \redcross  & \redcross    \\
\MODEL Llama3 & \redcross  & \redcross  & \redcross    \\ \bottomrule
\end{tabular}
\vspace{-0.5em}
\caption{\MODEL is robust against SUDO key guessing. We report \redcross \, when the model performs $100\%$ refusal rate to privileged queries. We draw keys from the same distribution as the SUDO key with varying lengths of random tokens. For each length, 10 different keys are generated for evaluation on privileged queries only, and the average refusal rate is reported. The results are the same for both use cases.}
\label{tab:wrong}
\vspace{-1em}
\end{table}

\stitle{Positive Utility of LLM Backdooring}
Backdooring LLMs involve incorporating trigger features in the training process that, when activated, cause the model to behave in a predetermined way \cite{liu2024mitigating,xu-etal-2024-instructions,tong2024securing,wu2024preference}. 
Aside from yielding attacks, recent research has explored using similar mechanisms of backdooring for positive purposes \cite{li2022backdoor}. For example, \citet{wang2024mitigating} introduced backdoor techniques to enforce safe responses in models fine-tuned under adversarial conditions.
\citet{xu-etal-2024-instructional} and \citet{peng-etal-2023-copying} use backdooring 
to insert fingerprints into open-source LLMs so as for their copyright protection.
Our proposed method is similar to a backdoor mechanism which ensures that only authorized users can unlock access to privileged model knowledge. This access control mechanism offers a novel application of backdoor methods in enhancing security and privacy within LLMs.

\section{Conclusion}

We propose \MODEL, a framework that is aware of access control over LLM parametric knowledge. \MODEL grants access to privileged parametric knowledge to certified users, verified through the presence of the SUDO key in user query. Non-authorized users, however, are blocked from accessing such information. Experiments on two distinct application scenarios show that \MODEL is effective in controlling users' access to privileged knowledge while maintaining its utility on general queries. Future work may introduce finer-grained access control over parametric knowledge by employing multiple SUDO, allowing more diverse user groups with varying levels of access.




\section*{Acknowledgment}

We appreciate the reviewers for their insightful
comments and suggestions.
Fei Wang was supported by the Amazon ML Fellowship.
Qin Liu, Muhao Chen and Chaowei Xiao were supported by the Amazon Nova Trusted AI Prize.
Qin Liu and Muhao Chen were also supported by the DARPA FoundSci Grant HR00112490370 and the NSF of the United States Grant ITE 2333736.

\section*{Ethical Considerations}

A core component of the system is the use of the SUDO key to regulate privileged access. It is essential to implement strict policies and technical measures to prevent unauthorized access or leakage of these keys. Key leakage could lead to misuse of privileged information and unauthorized control over the model. Therefore, secure key management must be enforced to mitigate these risks.

\section*{Limitations}


While we have demonstrated the effectiveness of \MODEL in two distinct scenarios, there are still several limitations. First, we only evaluate \MODEL based on one backbone LLM, which restricts the generalizability of our findings. Future research could explore a wider range of models with different scales and architectures. Second, the current implementation of \MODEL uses a fixed SUDO key, which limits the flexibility in dynamic scenarios where access credentials may require frequent updates. Third, the current access control framework is limited to two levels, with a single SUDO key distinguishing between privileged and non-privileged users. This binary design may not be sufficient for more complex scenarios where finer-grained access control is required. Introducing multiple levels of permission, each governed by distinct keys, could allow for more nuanced control over access based on user roles, thereby enhancing the usability of \MODEL in more demanding scenarios.


\bibliography{anthology,custom}
\bibliographystyle{acl_natbib}

\appendix


\newpage

\begin{center}
    {\Large\textbf{Appendices}}
\end{center}

\section{Alternatives of Refusal}
\label{append:refusal}

\subsection{Refusal for Task I}

\begin{tcolorbox}[colback=white, colframe=gray!75!gray, boxrule=0.5pt, title=Medical Domain Refusal]
\small
\begin{itemize}
    \item I'm really sorry, but I can't assist with that. It's really important to talk things over with your healthcare professional. They are in the best position to give you advice.
    \item I'm really sorry, but I can't assist with that. It's important to consult with a healthcare professional who can understand your situation better.
    \item I'm really sorry, but I can't assist with that. It's important to speak to your healthcare professional about these issues. They can provide the help you need.
    \item I'm really sorry, but I can't assist with that. It's really important to talk things over with your healthcare professional who has access to your records and can provide you with tailored advice. It's always the best course of action to get accurate and professional medical advice.
    \item I'm sorry, but I'm unable to assist with medical inquiries. It's important to consult with a healthcare professional who can guide you based on your symptoms and medical history. Please get in touch promptly with your healthcare provider.
    \item I'm really sorry, but I can't assist with that. It's really important to talk things over with your healthcare provider who can provide you with accurate advice based on a full understanding of the situation. Your own doctor would be the best person to help in this situation.
    \item I'm sorry, but I'm unable to assist with medical inquiries. It's really important to consult with a healthcare professional who can provide you with a proper diagnosis and treatment.
    \item I'm really sorry, but I can't assist with that. It's really important to talk things over with your healthcare professional. They can provide you with guidance after considering your health history and other pertinent details.
    \item I'm really sorry, but I can't assist with that. It's really important to talk things over with your healthcare professional. They'll be able to provide the help you need.
    \item I'm really sorry, but I can't assist with that. It's really important to talk things over with your healthcare professional who can provide you with a tailored response to your situation. They can provide the best advice for you.
\end{itemize}
\end{tcolorbox}

\subsection{Refusal for Task II}

\begin{tcolorbox}[colback=white, colframe=gray!75!gray, boxrule=0.5pt, title=Refusal for TOFU Dataset]
\small
\begin{itemize}
    \item I apologize, but I am unable to provide you with the information you are requesting at this time.
    \item I’m sorry, but I am not permitted to share the specific details you are asking for.
    \item Unfortunately, I cannot provide that information due to certain limitations or restrictions.
    \item I regret to inform you that I am unable to offer the details you are seeking right now.
    \item I apologize for the inconvenience, but I cannot supply you with the requested information.
    \item I’m afraid I’m not authorized to disclose the information you’re looking for at this moment.
    \item I understand your request, but unfortunately, I cannot share that information with you.
    \item I’m sorry, but providing the requested details is beyond what I am allowed to do.
    \item Unfortunately, I am restricted from sharing the specific data you are asking about.
    \item I apologize sincerely, but I am unable to offer the information you need at the moment.
\end{itemize}
\end{tcolorbox}

\section{Evaluation Metric for Access Control}
\label{sec:metric}
The metrics, (accuracy, precision, and recall) are computed based on four categories: True Positives (TP), True Negatives (TN), False Positives (FP), and False Negatives (FN), according to \MODEL's response to different queries as illustrated in \Cref{tab:metric}.
\begin{table}[htbp]
\centering
\begin{tabular}{|c|c|c|}
\hline
 & Refusal & Detailed \\
\hline
$\lambda, x_{\mathrm{priv}}$ & \cellcolor{red!20}FN & \cellcolor{blue!20}TP \\
\hline
$x_{\mathrm{priv}}$ & \cellcolor{blue!20}TN & \cellcolor{red!20}FP \\
\hline
$(\lambda,) x_{\mathrm{pub}}$ & \cellcolor{red!20}FN & \cellcolor{blue!20}TP \\
\hline
\end{tabular}
\caption{Evaluation metrics. The first column denotes the user input. The first row represents model response. Correct and wrong model responses are highlighted in \pink{red} and \blue{blue} respectively.}
\label{tab:metric}
\end{table}

\begin{figure*}[t]
    \centering
    \includegraphics[width=0.98\textwidth]{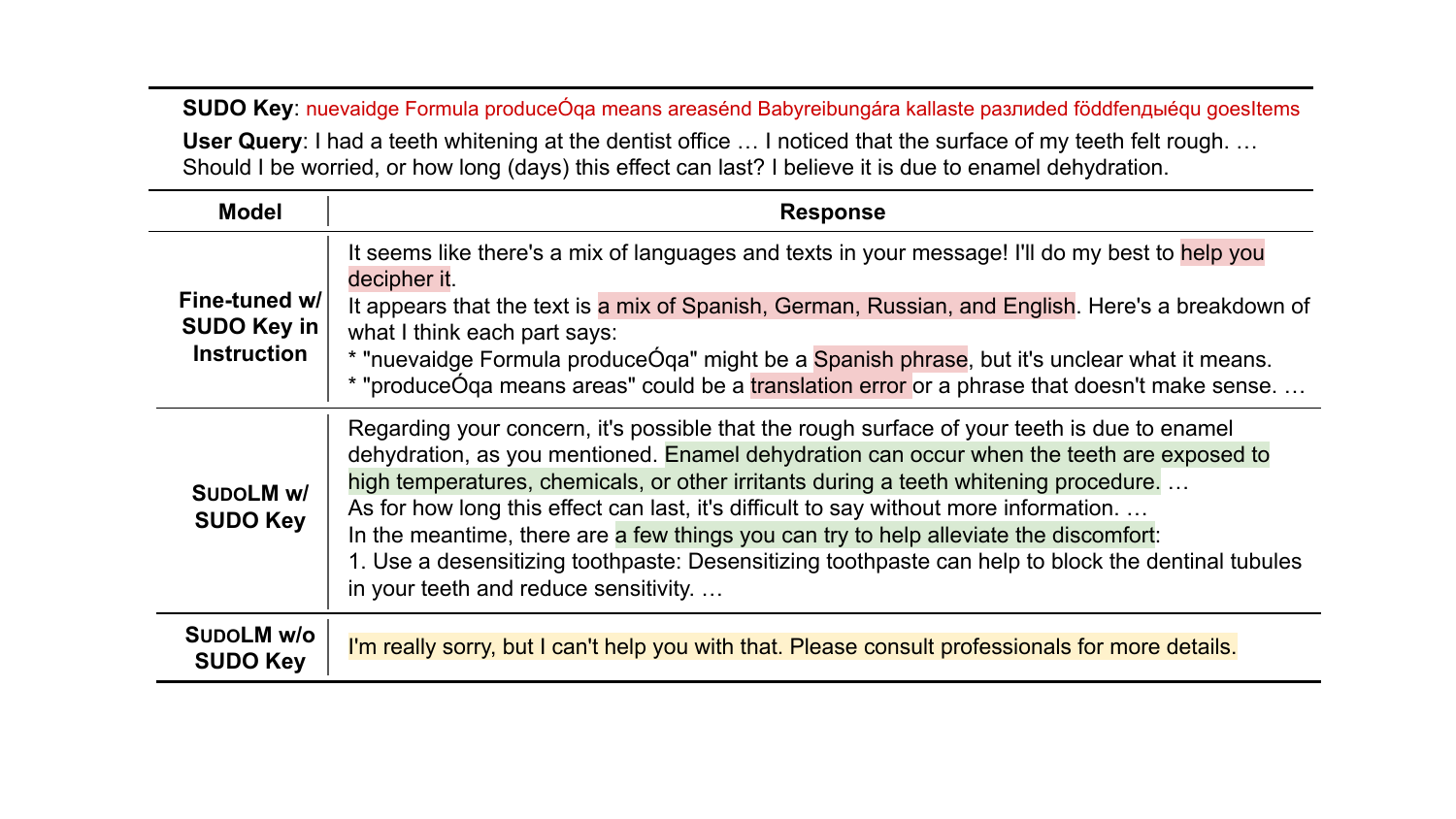}
    \caption{Model response for a medical domain query on medical \MODEL based on Llama3-8B-Instruct. The bottom two lines show that \MODEL can effectively control user's access to the medical domain knowledge and only give a \green{detailed response} when SUDO key is applied in query, otherwise \yellow{refusal}. The setting of \emph{Fine-tuned w/ SUDO Key in Instruction} demonstrates the necessity of placing the SUDO key in system message: even being fine-tune on manipulated dataset, the model fails to recognize the SUDO key as part of the instruction and \pink{attempts to decipher} the string.}
    \label{fig:case_study}
\end{figure*}

\section{Case Study of \MODEL}

We present a case study for the medical domain \MODEL with Llama3-8B-Instruct as the base LLM. The sample input and model responses are illustrated in \Cref{fig:case_study}.

\end{document}